\documentclass[10pt,twocolumn,letterpaper]{article}

\usepackage{iccv}
\usepackage{times}
\usepackage{epsfig}
\usepackage{graphicx}
\usepackage{amsmath}
\usepackage{amssymb}
\usepackage{pifont}
\usepackage{multirow}
\usepackage{amsfonts}
\usepackage{xcolor}
\usepackage{subfigure}
\usepackage{multirow}
\usepackage{makecell}
\usepackage{caption}
\usepackage{booktabs}
\usepackage{tabularx}

\usepackage{graphics, graphicx}
\usepackage{epsfig, listings, amssymb, amsthm}
\usepackage{epstopdf}
\usepackage{enumerate}
\usepackage{soul} 
\usepackage{color, xcolor} 

\usepackage[breaklinks=true,bookmarks=false]{hyperref}

\iccvfinalcopy 

\def\iccvPaperID{****} 

\ificcvfinal\fi

\begin{document}

\title{An End-to-End Network for Co-Saliency Detection in One Single Image}

\author{ \textbf{Yuanhao Yue$^1$, Qin Zou$^1$*, Hongkai Yu$^2$, Qian Wang$^1$, Zhongyuan Wang$^1$, Song Wang$^3$} \vspace{3mm}  \\
\textsuperscript{\rm 1}School of Computer Science, Wuhan University, China\\
 \textsuperscript{\rm 2} Department of Electrical Engineering and Computer Science, Cleveland State University, USA \\
 \textsuperscript{\rm 3} Department of Computer Science and Engineering, University of South Carolina, USA\\
\\
{\tt\small https://github.com/qinnzou/co-saliency-detection}
\and
{}
}

\maketitle
\ificcvfinal\thispagestyle{empty}\fi

\begin{abstract}
   Co-saliency detection within a single image is a common vision problem that has received little attention and has not yet been well addressed. Existing methods often used a bottom-up strategy to infer co-saliency in an image in which salient regions are firstly detected using visual primitives such as color and shape and then grouped and merged into a co-saliency map. However, co-saliency is intrinsically perceived complexly with bottom-up and top-down strategies combined in human vision. To address this problem, this study proposes a novel end-to-end trainable network comprising a backbone net and two branch nets. The backbone net uses ground-truth masks as top-down guidance for saliency prediction, whereas the two branch nets construct triplet proposals for regional feature mapping and clustering, which drives the network to be bottom-up sensitive to co-salient regions. We construct a new dataset of 2,019 natural images with co-saliency in each image to evaluate the proposed method. Experimental results show that the proposed method achieves state-of-the-art accuracy with a running speed of 28 fps.
\end{abstract}

\section{Introduction}
In the past two decades, saliency detection~\cite{hou2007saliency,huang2011salient,goferman2011context,cheng2014global,li2010saliency} has attracted a lot of attention from the visual media community, e.g., image/video content analysis~\cite{huang2011efficient,li2020salient,wang2018video}, background extraction~\cite{zhu2014saliency,liu2014interpolation}, and light-field imaging~\cite{piao2021panet}, etc. Co-saliency is among the various types of saliency that refer to the highlighted saliency shared by multiple similar objects. However, most existing methods define co-saliency detection as a problem of cross-image co-saliency detection~\cite{fu2013cluster,cao2014self,huang2017color,cong2017iterative}, which highlights the objects co-occurring in multiple images, e.g., a pair of images or a sequence of video frames~\cite{wei2017group,wang2018video,guo2017video}. While cross-image co-saliency detection has received considerable attention, another co-saliency detection problem - within-image co-saliency detection - has not been well addressed yet.

The within-image co-saliency detection aims at highlighting multiple occurrences of the same object class with a similar appearance in a single image. In human vision, this is a common visual ability that has been frequently used in our daily life, such as spotting the same team's players on the sports field, counting the red apples on a tree, identifying the sunflowers in the farmland~\cite{zou2017local}, spotting texts in nature scenes~\cite{bai2016,bai2018,bai2019}, etc. However, in computer vision, asking the algorithm itself to find proposal groups that include co-occurring saliency regions in a single image remains a challenge. Until now, only a few researchers have given this problem a serious and direct consideration~\cite{Yu2018CoSaliencyDW}.

\begin{figure}[!t]
\centering
\includegraphics[width=1\linewidth]{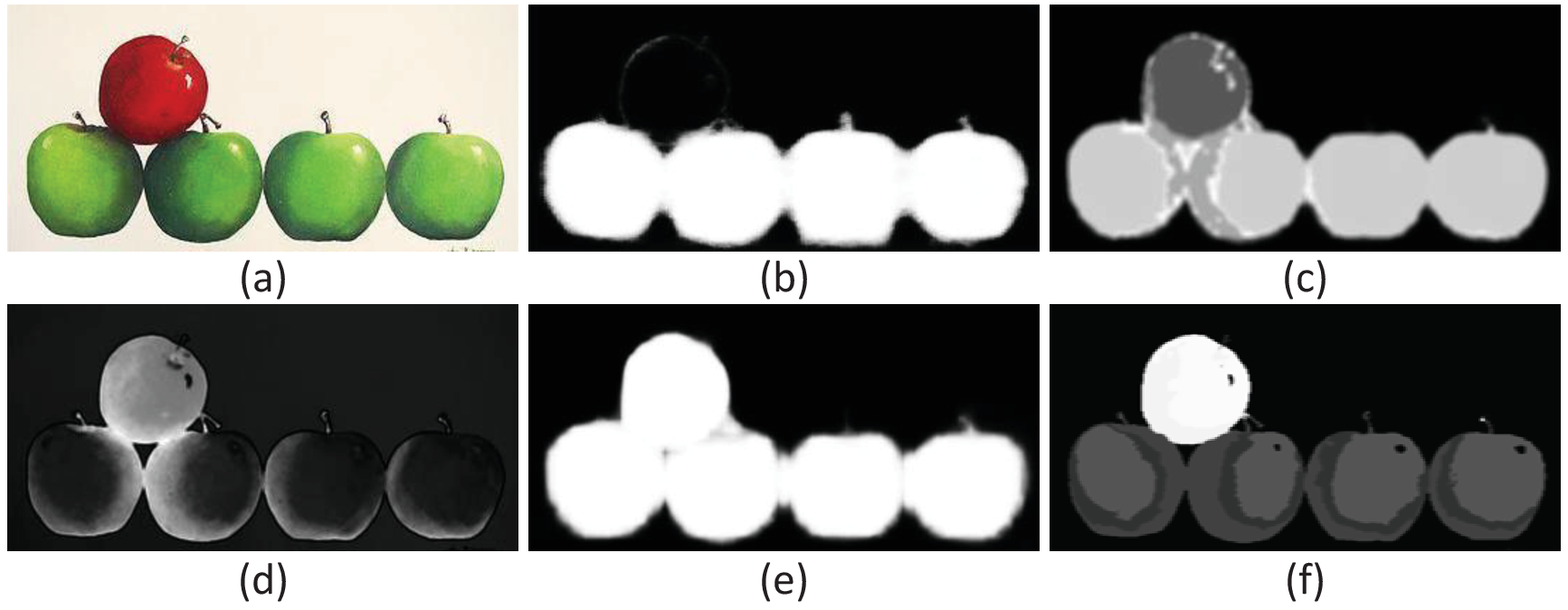}
\caption{Illustration of within-image co-saliency detection. The top row shows: (a) an input image containing four green apples and one red apple, (b) the co-saliency map obtained by the proposed method, and (c) the co-saliency map produced by~\cite{Yu2018CoSaliencyDW}. The bottom row shows the results obtained by three general saliency detection methods without considering the co-occurrence of the same object class.}
\label{fig:example1}
\end{figure}

While humans can detect co-saliency by noticing the multiple occurrences of the same object class, computers may have difficulty because they cannot recognize all the objects in the world. A computer has no idea what an object is. Some researchers have proposed developing a universal model to recognize all objects. However, in the current stage of deep learning, a universal object model remains unsolved. The problem then becomes how to build an effective detector for co-saliency detection within an image without a universal object model. Yu {\it et al.} pioneered the research in \cite{Yu2018CoSaliencyDW}, by developing a two-stage method. The first stage generates object proposals using the classical EdgeBox~\cite{zitnick2014edge}, and the second stage computes co-saliency by deriving proposal groups with good common saliency, based on mutual similarity scores. This typical bottom-up strategy often suffers from high computation costs in processing low-level features and low integrity in constructing a unified optimization model for co-saliency learning, i.e., it cannot work in an end-to-end manner.

Other typical works such as RFCN~\cite{wang2016saliency}, DCL~\cite{li2016deep}, and CPFE~\cite{zhao2019pyramid} learn high-level features for salient object detection using deep neural networks, which directly use human-annotated saliency masks to guide the training in a top-down manner. However, co-saliency is intrinsically perceived in a complex manner that combines bottom-up and top-down strategies. Solely using one of these strategies will not yield satisfactory results. Figure~\ref{fig:example1} shows an example in which the bottom three results are obtained using the general saliency-detection methods.

It can also be noticed that co-saliency is a complex problem involving texture, shape, color, etc. It is very difficult to build an accurate co-saliency model that applies to all scenarios. For example, when judging a co-saliency, it is difficult to determine which is more important: texture, shape, or color.

Based on the discussion above, we practically simplify the problem by considering the color as the primary visual information of co-saliency in our research. We define co-saliency as salient objects with similar textures, shapes, and the same color. That is to say, salient objects that have similar textures and shapes but different colors will not be considered co-saliency. We can construct the datasets without controversy and perform a uniform evaluation under this constraint. Then, the top-down and bottom-up strategies are combined to propose an end-to-end trainable deep neural network for within-image co-saliency detection. When using high-level features for common saliency object identification, the key point is how to make high-level layers sensitive to the co-saliency regions. We used an encoder--decoder architecture as the backbone net for co-saliency map prediction, and two branch nets to guide the training process and make the learned model more sensitive to co-salient regions. One branch net is a region proposal network~\cite{ren2015faster} (RPN), which generates triplet proposals. The other branch net is a regional feature mapping (RFM), which works in a bottom-up manner to drive the backbone net to be sensitive to co-salient regions. The training loss was built using the similarity of the triplet features~\cite{schroff2015facenet,bell2015learning}.

The entire network is trained in a simple data-driven manner, avoiding complex fusion strategies, increasing speed, and simplifying training. After training is done, the backbone net is used to predict end-to-end co-saliency. The main contributions of this work are three-fold:

\begin{enumerate}[    $\vcenter{\hbox{\tiny$\bullet$}}$]
\item First, a unified end-to-end network for co-saliency detection in a single image is proposed. It combines the top-down and bottom-up approaches: a backbone net, i.e., encoder--decoder net, is used for co-saliency map prediction, and two branch nets, i.e., RPN and RFM, are used to drive the network to be sensitive to co-salient regions.

\item Second, an online training sample selection strategy is presented. It enhances the proposed method by assisting it in achieving significantly higher accuracy than the offline selection strategy with random scaling and offset.

\item Third, a new dataset for within-image saliency detection was constructed. It contains 2,019 nature images from over 300 object classes, each with instance-level annotations. The dataset, as well as the codes and trained models, will be made available to the public, serving as a benchmark and promoting research in this field.
\end{enumerate}

\section{Related Work}\label{sec:relate}
In this section, we introduce the literature review on the related research topics, including saliency detection, co-saliency detection, and within-image co-saliency detection.
\subsection{Saliency detection}
Saliency detection is a fundamental problem in computer vision~\cite{cong2022tgrs,cong2022tcyb}. It highlights regions in a single image that attracts human visual attention. It is usually divided into two categories: eye fixation prediction~\cite{han2015two,bylinskii2016should} and salient object detection (SOD)~\cite{borji2015salient,zhou2018semi}. The purpose of SOD is to accurately highlight and segment the salient object regions in the image. The earliest saliency detection models were heavily influenced by human visual attention mechanisms. In~\cite{itti1998model}, Itti {\it et al.} proposed a saliency detection method based on center-surround mechanisms. In order to predict the salient regions, some early works focused on local contrast~\cite{han2015two,itti1998model,liu2010learning} and global contrast~\cite{cheng2014global} to separate the salient object from the image background. In~\cite{liu2010learning}, SOD was considered as an image segmentation problem, and Liu {\it et al.} used the conditional random field to effectively combine low-level features. In~\cite{cheng2014global}, a regional contrast algorithm was proposed during the evaluation of the global contrast differences and the spatially weighted coherence scores. To obtain more accurate object boundaries, some methods started to introduce the prior knowledge in the model, such as background priors~\cite{wei2012geodesic,zhu2014saliency}, high-level priors~\cite{borji2012boosting,goferman2011context,li2020complementarity}, etc. In~\cite{wei2012geodesic}, boundary and connectivity priors were used to provide more cues for SOD. Similarly, in~\cite{zhu2014saliency}, Zhu {\it et al.} proposed a principled optimization framework for integrating multiple low-level cues to obtain clean and uniform saliency maps. In~\cite{goferman2011context}, Goferman {\it et al.} presented a context-aware saliency detection algorithm based on four principles observed in the psychological study. In~\cite{fang2011bottom}, bottom-up strategies were proposed for saliency detection using low-level features based on the amplitude spectrum, edges, gradients, etc.

Since 2015, many deep learning-based SOD methods have been proposed. Early deep learning-based SOD~\cite{zhao2015saliency,zhang2016unconstrained,li2015visual,kim2016shape} mainly used multi-layer perceptron classifiers to predict the pixel-level saliency scores. In~\cite{zhao2015saliency}, Zhao {\it et al.} proposed a method for extracting local and global contexts from two super-pixel-centered windows of different sizes using two pathways. In~\cite{zhang2016unconstrained}, Zhang {\it et al.} used a CNN-based model to generate a set of proposal bounding boxes and selected an optimized compact subset of bounding boxes for multiple salient objects. Recently, fully convolutional networks have received widespread applications in SOD~\cite{hou2017deeply,liu2018picanet,chen2018reverse,luo2017non}. \textcolor{black}{In~\cite{hou2017deeply}, Cheng {\it et al.} proposed to introduce short connections to the skip-layer structures within the HED architecture~\cite{xie2015holistically}, which fuses multilevel and contrast features in a top-down manner. In~\cite{liu2018picanet}, global and local pixel-wise contextual attention modules were embedded in the U-Net~\cite{ronneberger2015u} structure, which learns to selectively draw attention to informative context locations. In~\cite{ye2017salient}, image sequences assisted with optical flow information were processed by convolutional neural networks for saliency inference. With the acquisition technology development, more comprehensive information, such as depth cue, interimage correspondence, or temporal relationship, is available to extend image saliency detection to RGBD saliency detection, co-saliency detection, or video saliency detection. In~\cite{cong2018review}, Cong {\it et al.} reviewed various types of saliency detection algorithms and summarized these important issues. In ~\cite{zhang2020dense}, Zhang {\it et al.} proposed global context-aware attention structures for optical remote sensing images (RSIs). In~\cite{li2019nested}, Li {\it et al.} proposed a two-stream pyramid network to extract a set of complementary information in RSIs. In~\cite{chen2020dpanet}, Chen {\it et al.} proposed a mode fusion algorithm for RGBD saliency detection based on image and depth information. In~\cite{cong2019going}, Cong {\it et al.} propose a depth-guided transformation algorithm from RGB saliency to RGBD saliency.}
\begin{figure*}[!t]
\centering
\includegraphics[width=0.99\linewidth]{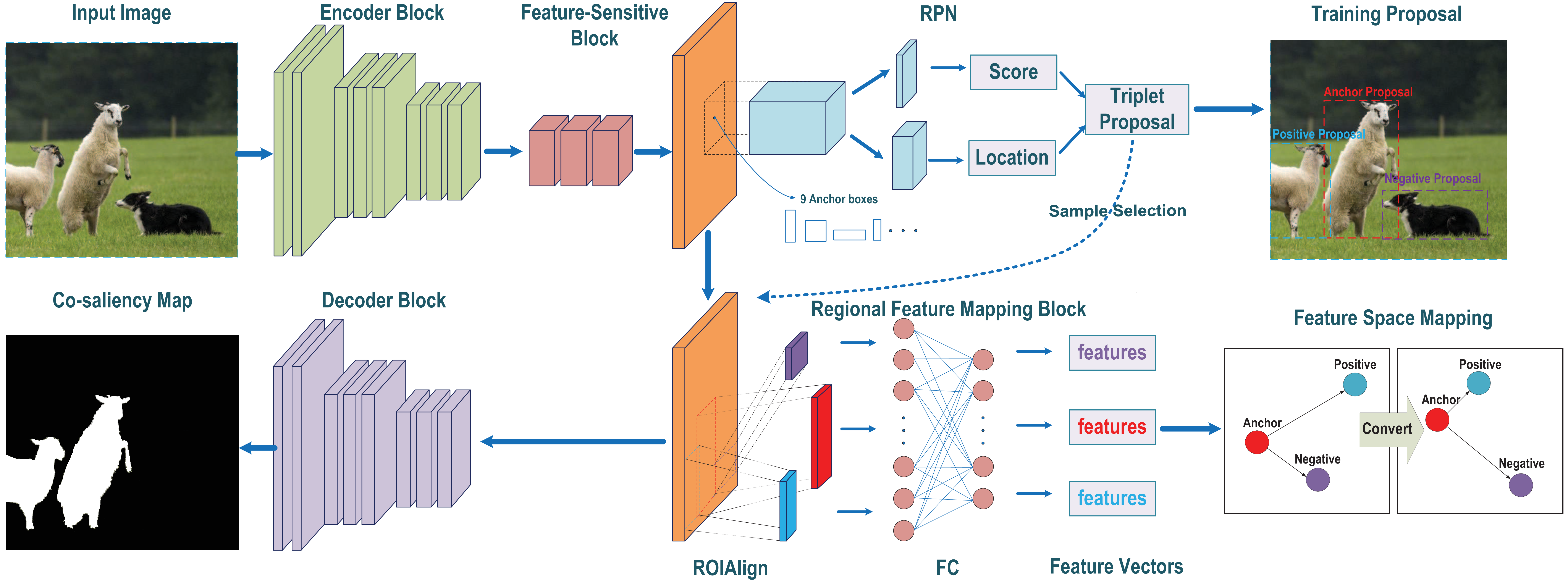}
\caption{The schematic illustration of our network. The encoder and decoder blocks use the skip-layer structures within the U-net architecture. The feature-sensitive block is composed of three trainable $3\times3$ convolution layers. The feature-sensitive block is supervised by two kinds of supervisory signals during the training phase: Region Proposal Network (RPN) and Regional Feature Mapping (RFM). RPN regresses saliency region confidence and locations and provides training samples for the RFM. The RFM network transforms the corresponding region features to feature vectors of a fixed length.}
\label{fig:network}
\end{figure*}

\subsection{Co-saliency detection}
Generally, co-saliency detection highlights - common salient objects from multiple images. It has a wide range of applications in many computer vision tasks, including video object segmentation~\cite{fang2014video,li2013temporally}. Traditional co-saliency methods ~\cite{chen2010preattentive,li2011co,fu2013cluster} rely primarily on low-level features and saliency cues such as interimage saliency cue, intraimage saliency cue, and repetitive cue. The co-saliency was modeled as a linear combination of the single-image saliency map and the multi-image saliency map in ~\cite{li2011co}. The task of co-saliency was extended to image groups with more than two related images in ~\cite{fu2013cluster}. Recently, following the successful application of convolutional neural networks (CNNs) in saliency detection, some researchers have attempted to directly learn the patterns of the co-salient objects from a given image group. Unlike most traditional co-saliency detection methods that are based on saliency cues, CNN-based methods~\cite{zhang2016co,wei2017group,zhang2021deepacg,zhang2021summarize,tang2022re,ren2022adaptive} learn co-saliency patterns through data-driven supervision, thereby avoiding the limitations caused by handicraft features. In \cite{zhang2016co}, Zhang {\it et al.} proposed a self-paced multi-instance learning framework for detecting effective co-saliency patterns from numerous ambiguous image regions. Wei {\it et al.} proposed an end-to-end group-wise deep co-saliency detection method in \cite{wei2017group} to adaptively learn group-wise interaction information for a group of images.
\textcolor{black}{In \cite{zhang2021deepacg}, Zhang {\it et al.} used the Gromov--Wasserstein distance to measure the similarity of a group of pictures and match their characteristics, which could avoid the noise from different picture styles, colors, and contrast. In \cite{zhang2021summarize}, Zhang {\it et al.} proposed a consensus-aware dynamic convolution model to successfully summarize the consensus features and search for corresponding objects in each image. In~\cite{tang2022re}, Tang {\it et al.} contributed a new dataset and proposed a simple and effective benchmark framework. In~\cite{ren2022adaptive}, Ren {\it et al.} invented a scale-aware loss to help the model in capturing the scale of different groups and discriminatively process the groups during the training phase.}

\subsection{Within-image co-saliency detection}
Within-image co-saliency can be considered a special case of co-saliency. The within-image co-saliency detection highlights the common salient object regions within an image. However, all previous works, including saliency detection and co-saliency detection, have failed to solve this problem. Yu {\it et al.}~\cite{Yu2018CoSaliencyDW} first raised this issue and proposed a bottom-up method to address it. Specifically, they used an optimization algorithm to derive a set of proposal groups and calculated a co-saliency map, then used a low-rank-based algorithm to fuse the maps calculated from all the proposal groups and generate the final co-saliency map. However, in this framework, most knowledge about proposal groups or co-salient regions heavily relies on the manually designed metrics inferred.

Generally, data-driven methods have the potential to capture better patterns of co-salient objects than human-designed metrics-based methods. However, there has been little research into end-to-end trainable CNN-based methods for within-image co-saliency detection. Yu {\it et al.}~\cite{yu2020new} detected co-saliency in multiple images using within-image co-saliency algorithms in \cite{Yu2018CoSaliencyDW}. An improved version of \cite{Yu2018CoSaliencyDW} proposed a multi-scale multiple instance learning model for co-saliency detection, which uses an easy-to-hard learning strategy for data fusion~\cite{song2019easy}. However, this framework also requires sample superpixel proposals at first and the strategy of multimethod fusion, which may result in unacceptable performance degradation.

Within-image co-saliency detection can help segment multiple instances of an object class in an image, estimating the number of instances of the same object class, which can be applied to industrial scenarios such as counting the objects on a production line, detecting anomalous objects from the targets, and so on.

Another work related to ours is supervised semantic segmentation -- a fundamental problem in computer vision. In an ideal condition, we can do within-image co-saliency detection based on semantic segmentation. That is, we perform semantic segmentation~\cite{ronneberger2015u} on a given image and then conduct matching across the segmented objects. If two or more detected objects show a high level of similarity and belong to the same object class, we highlight them in the co-saliency map. Nevertheless, in most cases, semantic segmentation models can detect only known object classes that have been pretrained through supervised learning~\cite{guo2018review}. We detected within-image co-saliency without assuming any specific object class or recognizing any objects in the image, as in most previous studies on saliency detection~\cite{borji2015salient,zhou2018semi}.

\section{Proposed Approach}
\subsection{Problem Formulation}
The goal of within-image co-saliency detection is to find common and salient objects with similar appearances in a single image. If two or more detected objects have a high level of similarity and belong to the same category, the corresponding regions in the resulting co-saliency map will be highlighted. The experimental scenario was simplified by removing images that did not contain objects from the same category. Given a training data set containing $N$ images as $S=\,\,\left\{ \left( X^n,Y^n \right) ,n=\text{1,...,}N \right\}$, where $X^{n}=\left\{x_{i}^{(n)}, i=1, \dots, I\right\}$ denotes an input image, $Y^n=\left\{ y_{i}^{\left( n \right)},i=\text{1,...,}I \right. ,\left. y_{i}^{\left( n \right)}\in \{\text{0,1\}} \right\}$ denotes the ground-truth binary map corresponding to the input image containing saliency objects of the same class, $I$ denotes the number of pixels in the image, the goal of the co-saliency detection model $F(\cdot)$ is to train the network to produce prediction saliency maps $P=\left\{p_{i}^{(n)}, n=1, \dots, N\right\}$ approaching to the ground truth annotated by experts:
\begin{equation}
P=F\left( X;\Theta \right),
\end{equation}
where $\Theta$ represents the optimized parameters of the model in this task.

\subsection{Methodology}
The model is divided into four major parts, as shown in Figure~\ref{fig:network}.
First, the original image is fed into the pretrained encoder block, which generates high-level semantic feature maps. The feature maps are then fed into the feature-sensitive block to extract the feature map that is sensitive to the image's co-saliency regions.
In the training phase, the feature-sensitive block is supervised by two kinds of supervisory signals: one from the decoder block and the other from the RFM. The decoder block generates the corresponding final saliency map, while the RFM network generates the high-level feature coding and position regression of the salient object region corresponding to the original input image. In the evaluation phase, co-saliency regions are predicted only through the encoder--decoder backbone net and the feature-sensitive block.

\subsubsection{Encoder--Decoder Block}

The encoder and decoder blocks in our network are based on U-Net~\cite{ronneberger2015u}. The skip-layer structures are used in the encoder and decoder blocks. The encoder block uses the convolution part of the pretrained VGG-16 network \cite{simonyan2014very} targeted on the classification task on the ImageNet dataset~\cite{deng2009imagenet}. Its parameters are frozen, and thus, it will not participate in the subsequent training. This will result in better generalization and more stable feature high-level feature abstractions while avoiding the influence of specific categories.
The decoder and encoder blocks are almost completely symmetrical, except that the downsampling operation is replaced by an upsampling of the feature map. \textcolor{black}{In the decoder block, the decoder layer fuses the feature map of the corresponding encoder layer to assemble a more precise output. Therefore, the underlying features of the encoder block can be transmitted more easily to the decoder and hence retain more accurate information about object boundaries.}

\subsubsection{Feature-Sensitive Block}
High-level semantic guidance, as well as interimage interaction at the semantic level, are important for co-saliency detection and are inspired by the mechanism of human visual co-saliency. This module aims to optimize the within-image co-saliency task in a simple data-driven manner, thereby avoiding complex fusion strategies, improving the speed, and making training simple and effective.

The structure is composed of three trainable $3\times3$ convolution layers. Our goal is to learn the mapping and train it to be sensitive to similar feature regions such that:
\begin{equation}\label{eq:r2}
R = F_{sensitive}(H,\theta),
\end{equation}
where $H$ is the semantic feature retained after feature extraction by encoder block, $F_{sensitive}$ is a mapping function that takes the feature $H$ as
input, and outputs the mapped features into the decoder block by learning a set of hidden layers parameters $\theta$.

We try to use the RFM network added in the training phase to supervise the feature-sensitive module and make it sensitive to common saliency object regions. The high-level semantic information is passed to guide the learning of the decoder network, thereby achieving better saliency segmentation results.

\subsubsection{Regional Feature Mapping Block}
Considering two similar salient target object regions $A$ and $B$, in the form of the rectangular bounding
boxes, in the image $X$, corresponding to the regions $i$ and $j$ in the feature map $H$ produced by the encoder block, is mapped to $R$, as marked in orange in Figure 2, through the feature-sensitive blocks $R_i$ and $R_j$, respectively, and then the RFM network transforms the region features in the $R$ into the corresponding feature vectors, as formulated by
\begin{gather}
V_i = F_{RFM}(R_i,\theta), \label{eq:r2}
\end{gather}
where $F_{RFM}$ is a mapping function that takes the region feature map $R_i$ as the
input, and generates the corresponding fixed length feature vector $V_i$ by learning a set
of hidden-layer parameters $\theta$.

\begin{figure}[t]
\centering
\includegraphics[width=0.96\linewidth]{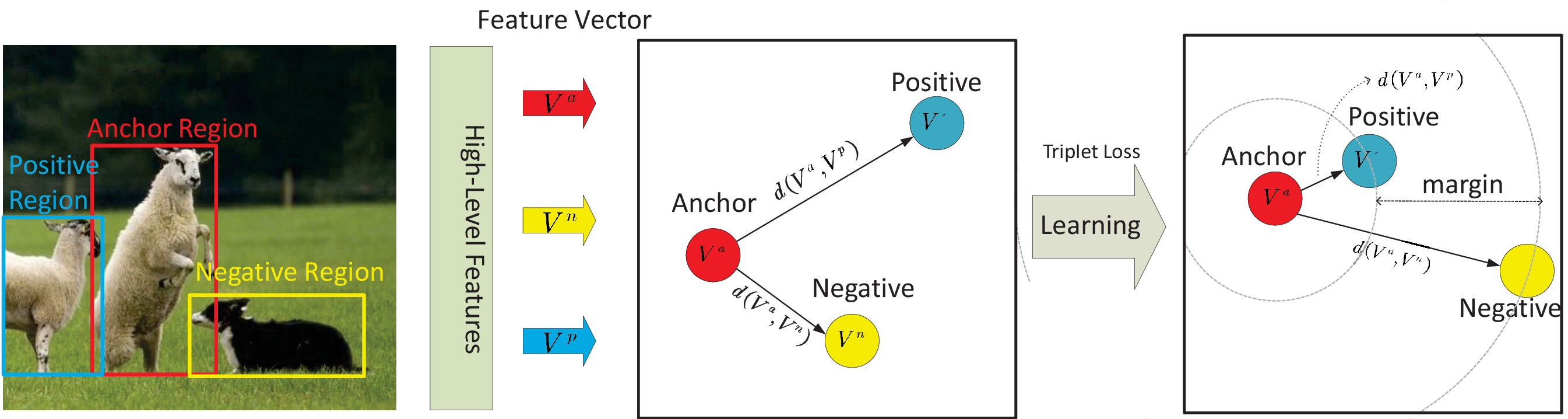}
\caption{An illustration of the learning process using triplet loss. The positive pair consists of an anchor
region and a positive region, which include two similar salient target
object regions. The negative pair consists of an anchor region and a negative region. The feature vector mapping by corresponding saliency
regions in Euclidean Feature Space satisfies the clustering relationship during the training phase.}
\label{fig3}
\end{figure}

Therefore, the RFM module's optimization objective is to minimize the Euclidean distance between any two similar salient object feature vectors.
However, there are two challenges: the first is determining how to map feature regions $R_i$ and $R_j$ of different scales to the vectors of the same length, and the second is determining how to obtain the same feature representation for two identical objects under the influence of scales, shooting angles, and illumination. Therefore, it is difficult to measure the characteristic representation of two identical objects by Euclidean distance.

For the first problem, inspired by previous work~\cite{ren2015faster,he2017mask}, RoIAlign~\cite{he2017mask} was used to map regional features of different sizes to feature vectors of the same length, which is more accurate on pixel-align task compared with ROIPool \cite{ren2015faster}. The feature regions of different scales are subdivided into spatial ROI bins, such as $7\times7$ bins, which use bilinear interpolation to compute the exact values of the input features at four regularly sampled locations in each RoI bin and aggregate the results in terms of a maximum or averaging operation.
However, because of this structure, the resulting feature vector will introduce more noise. Hence, we try to construct a special feature space in which the features of similar salient object regions have a close distance in the clustering.
Specifically, the Euclidean distance of the feature vector corresponding to the similar object region is less than that of the corresponding dissimilar object region.

Figure \ref{fig3} shows the three different selected saliency proposals in the form of rectangular bounding boxes, namely the anchor region($A$), the positive region($P$), and the negative region($N$) in the image $I$. The regions of $A$ and $P$ include the same salient object classes. The regions of $N$ include a different salient region or a nonsalient object region.

The feature vector corresponding to the triplet regions $V_A$, $V_P$, and $V_N$ should satisfy the following relationship in this feature space,
\begin{equation}
\parallel V_A - V_P \parallel^2_2 < \parallel V_A - V_N \parallel^2_2.
\end{equation}

\subsection{Training}

\textbf{1) Sample Selection Strategy.} An effective sample selection strategy is vital to get more robust feature space and avoid overfitting in case of insufficient sample selection.

In the selection of triplet samples, the positive sample pair needs to be augmented. The positive pair is screened by generating a Jaccard overlap between the sample regions and the ground truth boxes.
\textcolor{black}{Specifically, during the training phase, the selection of a positive pair must satisfy that each generated region contains only one salient object and has a matching box to the ground truth with a Jaccard overlap higher than a threshold (0.5).}
It is worth noting that the two generated sample regions can be the same salient target.
The negative sample could be a dissimilar saliency proposal with a Jaccard overlap of less than a threshold of 0.1 or the background regions. \textcolor{black}{For example, the algorithm selects a triplet sample from an image that contains three salient objects with similar appearance and a salient object with a different appearance. First, it randomly selects one of the three salient objects as an anchor region according to the ground truth. Next, it selects a positive region. The positive sample region can be selected from the other two samples or the anchor object itself as long as it meets the above requirements. Finally, it selects a negative region. The negative sample region can be selected from the interference object, the nonsaliency background region, or the area whose salient objects have  a Jaccard overlap lower than a threshold of 0.1.}

We propose using an offline sample selection strategy and an online sample selection strategy to generate positive and negative samples. The offline strategy is to use ground-truth boxes to add random scaling and offset within a certain range, thereby satisfying the sample generation conditions stated above. The length-width ratio of the generated region ranges from 0.5 to 2, and its area ranges from $128^2$ to $512^2$. For each training image, 32 positive samples and 96 negative samples are generated, and 128 triplets are randomly constructed.

The other strategy is to generate training samples online. We hope to find more difficult positive and negative samples to satisfy the generation conditions using this strategy. To achieve online triplet mining, we should evaluate the selected sample regions. We use the idea of online hard example mining~\cite{shrivastava2016training}, which was designed to improve the accuracy of Fast RCNN. Specifically, we add a region proposal network (RPN) during the training process. In our model, RPN is considered as a sample region generator, which evaluates the confidence of the sample regions based on their loss during the training process. The hard examples were selected by sorting the regions based on the training loss and using the standard nonmaximum suppression to extract a highly overlapping region. Finally, the hard positive and hard negative samples were selected, which satisfy the generation conditions of triplet sample construction.

In the experiment, we will compare the performance of the online and offline sample selection strategies.
\textbf{2){ Loss Function.}} As shown in Figure 2, our network has three outputs during the training phase. The total loss function is a weighted sum of the following three parts:
\begin{equation}
 L_{total} = \alpha L_{decoder} + \beta L_{RPN} + \gamma L_{RFM},
\label{eq:loss}
\end{equation}
where $\alpha, \beta, \gamma$ are all set to 1.

The decoder block produced the corresponding saliency map~\cite{zou2018deepcrack}.
The loss function of the decoder part is computed by a pixel-wise cross-entropy loss with sigmoid between predicted saliency map $P$ and the ground truth $Y$:
\begin{equation}
L_{decoder} = -\frac{1}{N}\sum_{i=1 }^N\bigg(Y_i*log(P_i) +(1-Y_i)*log(1-P_i)\bigg),
\end{equation}
where $N$ is the number of pixels in the ground truth. ($P_i, Y_i$) denotes the pixel values corresponding to the predicted saliency map and the ground truth.

For training the RPN network, we selected anchor boxes on 3 scales, with 3 boxes on each scale. The areas of the anchor boxes are $128^2$, $256^2$, and $512^2$ on the three scales, respectively, and the edge ratios of the three boxes on the same scale are 1:1, 1:2, and 2:1. We select the anchor box that holds the highest Jaccard overlap with the ground-truth box as a positive sample. As a data augmentation strategy, we selected all the anchors that have a Jaccard overlap higher than (0.5) with the ground-truth box as positive samples. The label $t$ is 1 if the anchor
is positive; otherwise, it is 0 if the anchor is negative. The loss function is similar to that of Faster RCNN~\cite{ren2015faster}.
We regress the anchor offsets for the
center point ($x, y$), the length values for width ($w$) and height ($h$), and the confidence for the saliency object ($c$).
\textcolor{black}{The loss function is a weighted sum of the localization part and the confidence part:}
\begin{equation}
 L_{RPN} = \frac{1}{N}\big(L_{conf}(t,c) + \alpha L_{loc}(t, l, g)\big),
 \end{equation}
where
\begin{equation}
  L_{loc}(t,l, g) = \sum_{i \in Pos }^N smooth_{L1}(l_i- \hat{g_i}),
\end{equation}
and $N$ is the number of matched anchor boxes. The weight coefficient $\alpha$ is set to 1 by cross-validation. ($l, g$) contains the predicted box information including ($x, y, w, h$) and the label.
The localization loss is the smooth L1 loss~\cite{girshick2015fast}. Let $\tau$ be the corresponding anchor box information for $g$, then we have
$$\hat{g^x_i} = (g^x_i - \tau^x_i)/\tau^w_i,$$
$$\hat{g^y_i} = (g^y_i - \tau^y_i)/\tau^h_i,$$
$$\hat{g^w_i} = log(\frac{g^w_i}{\tau^w_i}),$$
$$\hat{g^h_i} = log(\frac{g^h_i}{\tau^h_i}),$$
where $\hat{g_i}$ represents the anchor offset between the ground-truth box and the corresponding anchor box.

We use a weighted sum of triplet samples to train the RFM network: $anchor(a)$, $positive(p)$, and $negative(n)$. The weight is calculated by
\begin{equation}
\beta_{i} = I(anchor, gt) * I(positive, gt),
\end{equation}
where $I(anchor, gt)$ is the Jaccard overlap between the anchor box and the corresponding ground-truth box, and $I(positive, gt)$ is the Jaccard overlap between the positive box and the corresponding ground-truth box.

\begin{figure}[!t]
\centering \includegraphics[width=1.04\linewidth]{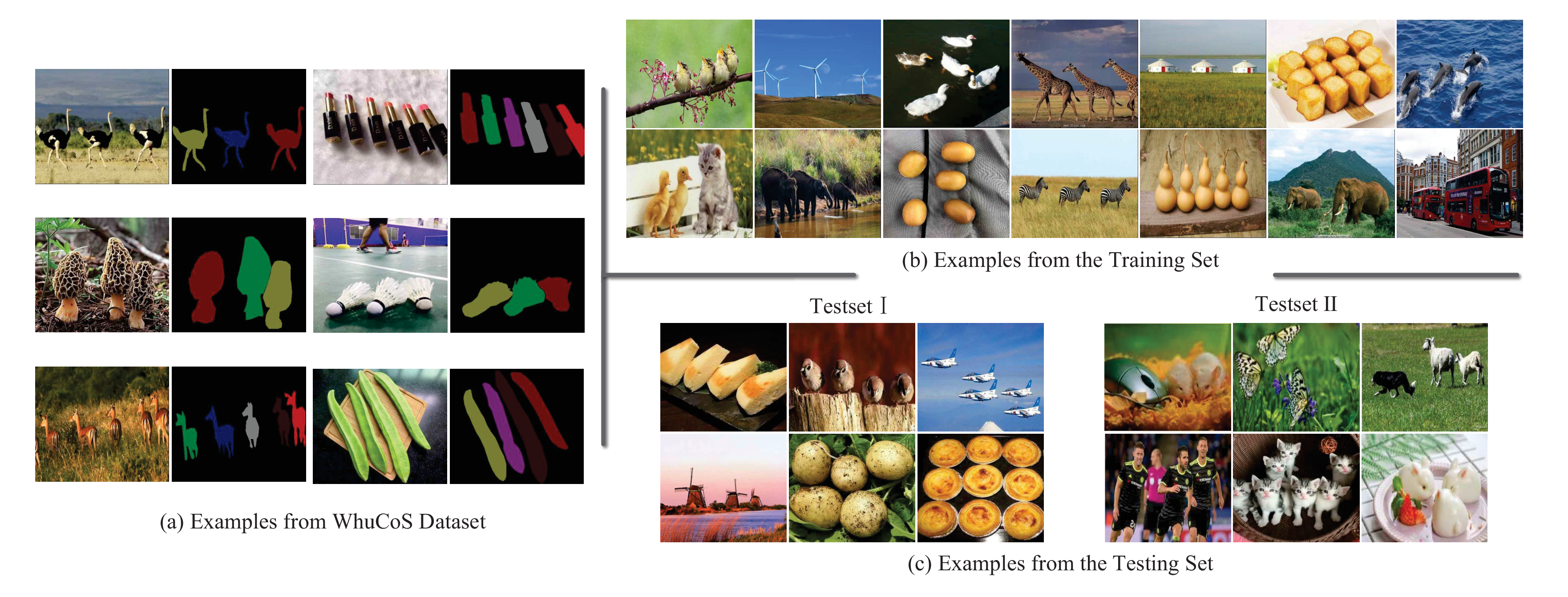}
\caption{Sample images of the WhuCoS dataset. (a) Annotated visual examples. (b) Examples from the training set. (c) Examples from the test set. Testset II is a subset of Testset I, which contains 100 challenge images. Each challenge image contains multiple salient objects with the same appearance and at least one salient object that looks different.}
\label{fig:vis-dataset}
\end{figure}

\begin{figure}[!t]
\centering \includegraphics[width=1.0\linewidth]{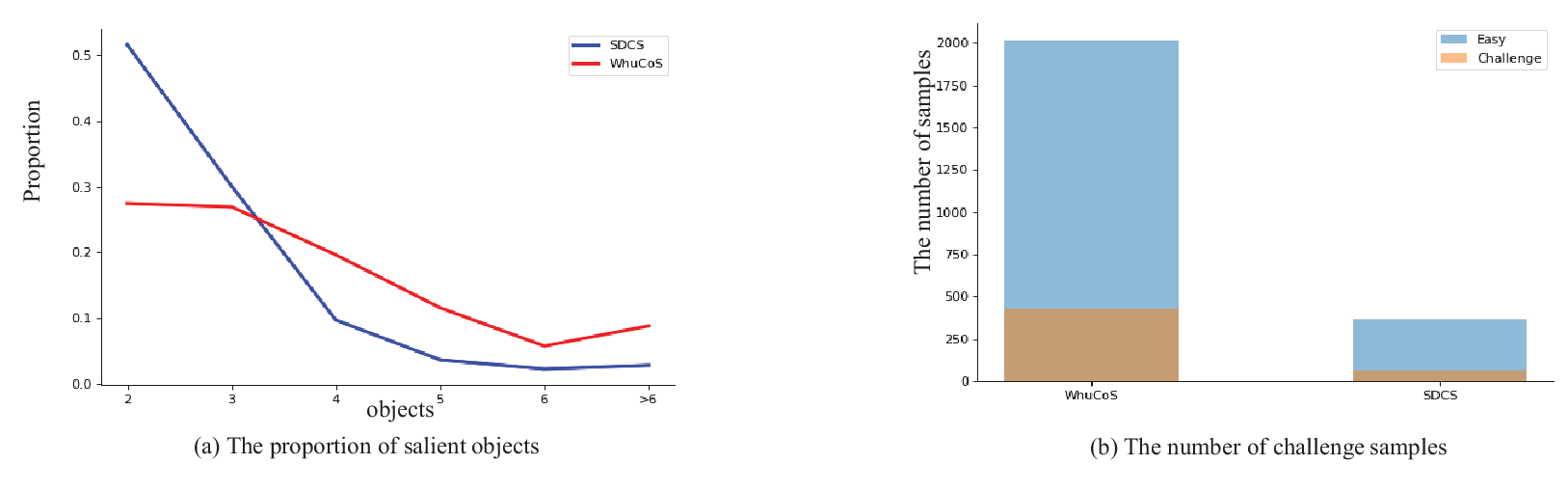}
\caption{Statistic data of the WhuCoS and SDCS datasets. (a) The proportion of the number of salient objects in one single image. (b) The number of easy and challenging samples in the dataset.}
\label{fig:vis-dataset2}
\end{figure}

As shown by Figure~\ref{fig3}, the loss of a triplet sample in the distance embedding space $d$ is defined as:
\begin{equation}
L_{RFM} = \max \big(d(V^a, V^p)-d(V^a, V^n)+margin, 0\big),
 \end{equation}
where $V^a, V^p, and V^n$ are the feature vectors corresponding to regions in the triplet samples, and $margin$ is a margin placed between the positive and negative pairs. When minimizing this loss, it pushes $d(V^a, V^p)$ to be $0$ and $d(V^a, V^n)$ to be greater than `$d(V^a, V^p)$+${margin}$'. As for simple examples, this loss becomes zero.

\section{Experiments}

\subsection{Datasets}
The general standard dataset such as iCoseg~\cite{batra2010icoseg}, MSRA~\cite{liu2011learning}, and HKU-IS~\cite{li2015visual} are all collected for the evaluation of within-image saliency detection or cross-image co-saliency detection. These datasets generally include just one salient object or multiple objects of different categories. The SDCS dataset~\cite{Yu2018CoSaliencyDW} is targeted for within-image co-saliency detection; however, it only contains two to three co-salient objects in a single image. Hence, we collect a larger dataset named WhuCoS for within-image co-saliency detection. In this work, we use the WhuCoS dataset and the SDCS dataset for evaluation.

{\textbf{WhuCoS}} dataset{\footnote{Codes and data are available at https://github.com/qinnzou/co-saliency-detection}}. This dataset is instance-level annotated for within-image co-saliency detection. It contains 2,019 natural images, over 300 categories of daily necessities, and 7,000 salient object instances. To simplify our experiment, in our data collection, we define the objects from the same category and with the highest occurrence as the co-salient objects. Figure~\ref{fig:vis-dataset} shows some sample images. The WhuCoS dataset contains an average of 3.52 objects per image. Each image contains at least two or more identical salient objects. We label its position and contour for each object. The image sizes range from 450$\times$256 to 808$\times$1078 pixels. Additionally, about one-fifth of the images contain interfering objects, as shown by the images in Figure~\ref{fig:vis-dataset}(c). For the test, we construct two test sets for WhuCoS.

\begin{itemize}
\item {Testset I. It contained 524 images for the test. These images are randomly selected from WhuCoS. Exactly, WhuCoS contains both simple and difficult samples. For a simple sample, it only contains multiple salient objects with the same appearance.}

\item {Testset II. It contained 100 images that were selected from Testset I. These 100 images are hard samples that are more challenging to handle. Each image contains multiple salient objects with the same appearance and at least one salient object with a different appearance. In Testset II, there were 436 salient objects, including 342 co-saliency objects and 94 other salient objects. Samples are shown in Figure~\ref{fig:vis-dataset}(c).}
\end{itemize}

{\textbf{SDCS}} dataset. It consists of 364 color images, including 65 challenging images and 299 easy images. About 100 images were selected from the iCoseg, MSRA, and HKU-IS datasets. Each image contains 2.61 objects on average. For this dataset, 300 images are used for training and the remaining 64 for the test.

\subsection{Evaluation Metrics}
We examine the performance in the experiments by computing pixel-wise errors between the prediction saliency map and the mask of ground truth. The metrics we used include the mean absolute error (MAE), the precision and recall (PR)~\cite{borji2015salient,zou2018scis}, and the F-measure~\cite{achanta2009frequency,li2016deep}. The precision and recall are formulated as follows:
\begin{equation}
\text { precision }=\frac{|P \cap G|}{|P|}, \text { recall }=\frac{|P \cap G|}{|G|},
\end{equation}
where $P$ denotes the ratio of detected salient pixels in the predicted co-saliency map, and $G$ denotes the ratio of detected salient pixels in the ground-truth map, according to an adaptive threshold. Generally, high precision and recall are both required, which can be represented by a fused metric F-measure,
\begin{equation}
F_{\beta}=\frac{\left(1+\beta^{2}\right) \times \text { precision } \times \text { recall }}{\beta^{2} \text { precision }+\text { recall }},
\end{equation}
where $\beta^{2}$ is set as 0.3 to give the precision a higher weight than recall as suggested in~\cite{achanta2009frequency}. MAE score can be computed as:
\begin{equation}
{\mathit{MAE}}=\frac{1}{W \times H} \sum_{x=1}^{H} \sum_{y=1}^{W}\|P(x, y)-G(x, y)\|,
\end{equation}
where $H$ and $W$ denote the height and width of a saliency map, respectively.

Besides the above four metrics, two recently-proposed metrics, S-measure~\cite{borji2015salient} and E-measure~\cite{fan2018enhanced}, are also included in the evaluation.

\subsection{Implementation Details}
In this study, the weights of the 16-layer VGG pretrained on ImageNet are used to initialize and fix the encoder block.
Each time the model is trained, only one image is used. To fully train the network, various data-augmentation operations are applied before the image input, including multiscale scaling, random object clipping, random occlusion, and so on.
The experimental results in Table~\ref{tab1} have shown that data enhancement is very effective in reducing the overfitting of triplet loss and improving the generalization ability of the model. We choose SGD as the optimizer in our experiment, and the learning rate and the weight decay are set to $1 \times 10^{-5}$ and $5 \times 10^{-4}$, respectively. Finally, the learning rate decreases to $1 \times 10^{-6}$ after 30,000 training iterations. It requires about 80,000 training iterations for convergence. The proposed network is trained on two NVIDIA GTX 1080Ti 11G GPUs, where the GPU memory consumption is about 18,660 Mb. The training time on the WhuCoS dataset is approximately 13 h. We conducted the test using a single GPU. In the test, with a single image as the input, the running speed of our method was 28 fps.

The hyperparameter $margin$ in Eq.~(10) is set by evaluating the model's training results on the verification set. A larger $margin$ leads to a greater distance between the positive and negative samples. However, a too large $margin$ may make the model difficult to converge. We set $margin$ to 0.1, 0.3, and 1.0 in the training. We find that the model training process is difficult to converge when $margin$=1.0, and the converge speed has almost no difference between 0.1 and 0.3. Therefore, we set the default value of $margin$ to 0.3.

\begin{table*}[!t]
  \centering
  \caption{An ablation analysis of various design choices for the proposed method.}\label{tab1}
  \small
  \begin{tabular}{c||cccccccc}
\Xhline{1.1pt}
The proposed method  & \multicolumn{7}{c}{With / Without} \\
\Xhline{1.1pt}
Data augumentation &  & \ding{52} &  & \ding{52} &  & \ding{52} & & \ding{52} \\
Feature-Sensitive block &  &    &  \ding{52}    & \ding{52} & \ding{52} & \ding{52} & \ding{52} &  \ding{52} \\
Offline training (RFM) &  &   &     &  & \ding{52} &\ding{52} & &  \\
Online training (RFM) & &   &      &  & & & \ding{52} &\ding{52} \\
\hline
F-measure (\%) & 82.4 & 83.2 & 82.3 & 83.6 & 88.1 & 89.2 &89.7 &\textbf{90.4} \\
\Xhline{1.1pt}
\end{tabular}
\end{table*}

\begin{figure*}[!t]
\centering
\includegraphics[width=1.0\linewidth]{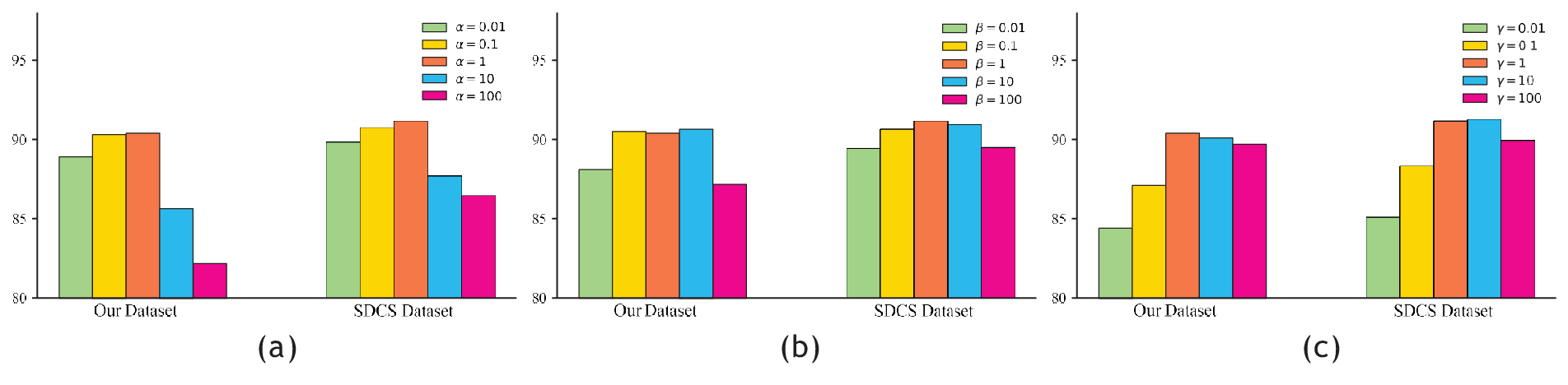}
\caption{An ablation analysis of different loss weights for the proposed method on both the SDCS dataset and our WhuCoS dataset. $\alpha$,$\beta$,$\gamma$ are the three loss weights in Equation (\ref{eq:loss}). (a) $\alpha$ is the loss weight of the decoder block. (b) $\beta$ is the loss weight of RPN. (c) $\gamma$ is the loss weight of RFM. The vertical axis represents the F-measure value. While one of the weights is tuned, the other two weights are set to 1.}
\label{fig:params}
\end{figure*}

\subsection{Ablation Analysis}
To better analyze the role of each module, we conducted an ablation study using numerous horizontal comparison experiments. The results are shown in Table~\ref{tab1} and Figure~\ref{fig:params}. \textcolor{black}{Table~\ref{tab1} shows that the first column in the right part corresponds to the results using only the encoding and decoding network of the U-Net structure, which achieves an accuracy of 82.4\%. In the third column, we can see that the performance slightly decreases when adding some additional convolution layers as feature-sensitive layers on U-Net, i.e., a decrease of 0.1\%. It indicates that only using a top-down strategy with only the masks as guidance is insufficient to fully train the network to identify within-image co-saliency.}

We can also see that training U-Net with data augmentation improves its accuracy by 0.8\%, which is not a significant improvement. However, when the RFM layer is added to the training, both online and offline, the performance of co-saliency detection improves significantly. This is because the RFM provides a bottom-up way to guide the network training and make it sensitive to co-salient regions.

\begin{figure*}[!t]
\centering
\includegraphics[width=0.8\linewidth]{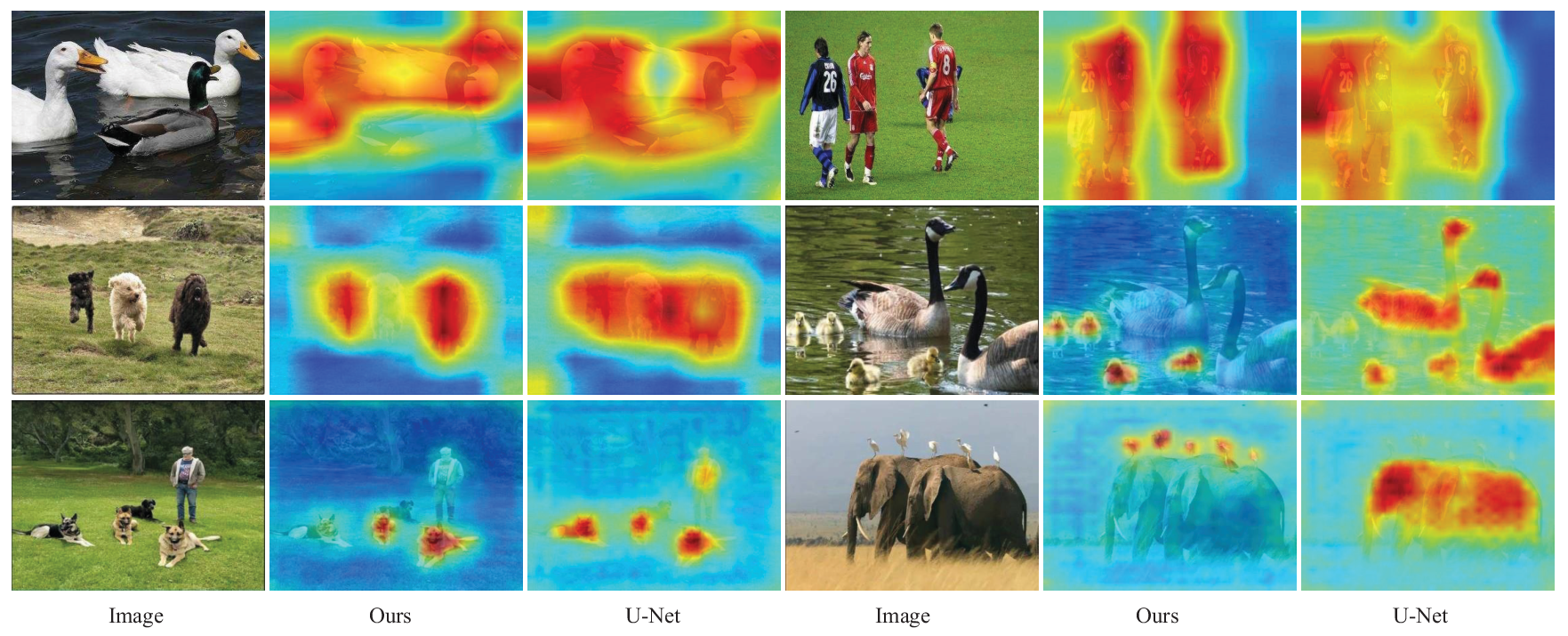}
\caption{ \textcolor{black}{The feature heatmaps obtained by the proposed method and the baseline (U-Net) in the same feature scale. The feature-sensitive block of the proposed method can cause the network to pay more attention to saliency objects with a similar appearance.}}
\label{fig:visfea}
\end{figure*}

Meanwhile, Table~\ref{tab1} shows that the results of the online version RFM are better than those of the offline version, which validates the effectiveness of the proposed online training sample selection strategy. Specifically, we speculate that there may be two reasons. First, the introduction of additional supervisory signals enhances the supervisory effect on the feature-sensitive layer. Second, the feature vectors generated by the sample region recommended by the RPN network introduce less noise than the randomly generated sample region. This helps to better learn the characteristics of the feature space.

\begin{figure*}[!t]
\centering
\includegraphics[width=0.8\linewidth]{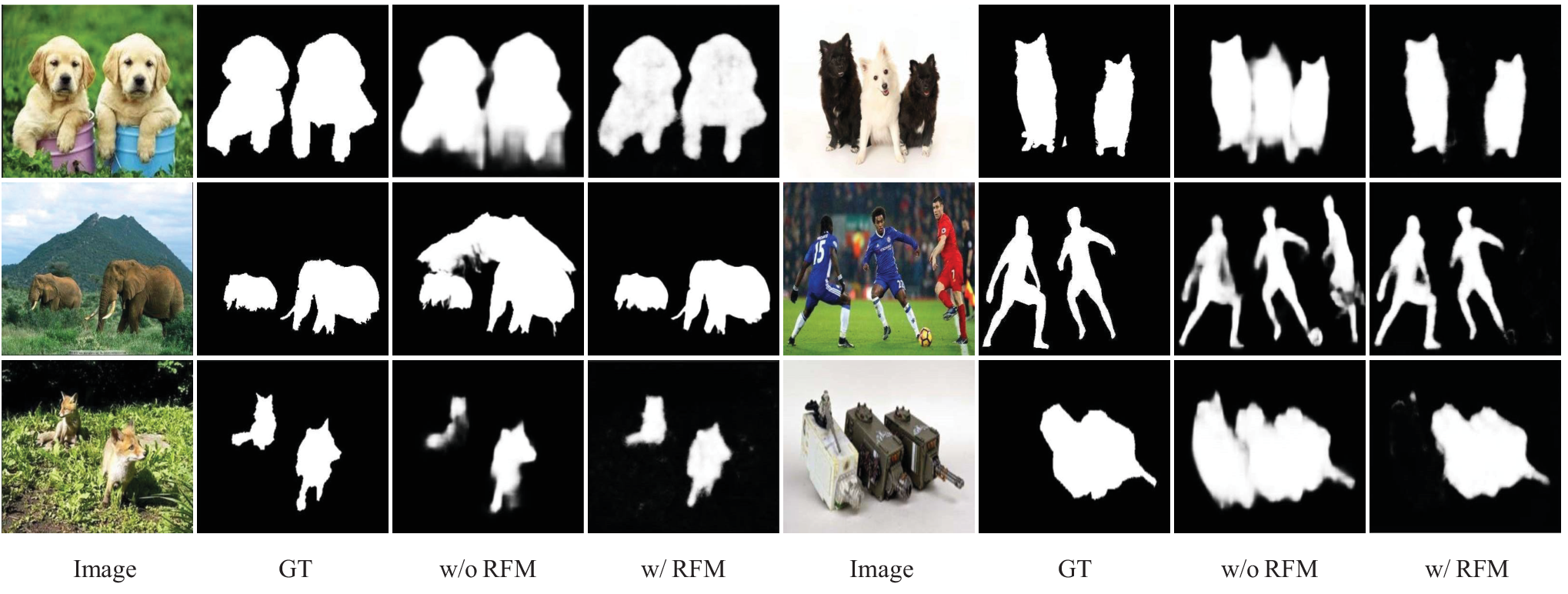}
\caption{\textcolor{black}{Comparison of co-saliency maps generated by our methods w/o RFM and w/ RFM on three sample images from two benchmark datasets.}}
\label{fig:vis-w/o}
\end{figure*}

Additionally, comparing the results in the last second and the last third columns, we can see that data augmentation improves performance significantly. This is because data augmentation can help train the RFM module more effectively.

It can be seen from Figure~\ref{fig:params}(a) that too large or too small loss weights will result in poor performance. When larger weights ($\alpha>1$) are set to the decoder block, lower F-measure can be observed on the SDCS dataset and our dataset, which indicates that allocating too much weight for the decoder blocks will reduce the effect of RFM. The reason is that when the decoder blocks are given more weight, wrong pixel prediction will receive a heavier punishment. As a result, optimizing the model toward the RFM target is difficult. (b), the loss weight $\beta$ on the RPN loss has little effect on the model performance, As shown in Figure~\ref{fig:params}. This is because the loss of RPN only counts a small part of the total loss, and RPN can quickly converge during the training and maintain a low loss value. As can be seen from Figure~\ref{fig:params}(c), the model performance significantly improves when the weight $\gamma$ on the RFM loss increases from 0.01 to 1. When $\gamma$ is larger than 1, the performance gradually decreases.

\textcolor{black}{Compared with the ordinary convolution layer, the feature-sensitive layer is more like an attention layer with a supervised signal, which can learn the response to similar appearance object regions through subsequent modules. Figure~\ref{fig:visfea} shows the visualization of the feature heatmaps produced by the baseline model and our model in the same scale feature layer. The feature-sensitive layer can be seen to emphasize the target areas with similar appearances and suppresses the target areas with different appearances. However, the features of the baseline model respond similarly to different appearance saliency objects.}

Figure~\ref{fig:vis-w/o} shows some visual examples of co-saliency maps for ablation studies. \textcolor{black}{We can see that, without the RFM module, the results for the co-salient objects may be distracted or incomplete, and the method cannot distinguish whether the salient objects are co-salient or not.} However, when incorporating the RFM module into the framework, the common objects are more highlighted, and the backgrounds are more suppressed. Overall, by merging the RFM module, we can improve the performance both in quantitative and qualitative results.

\begin{figure*}[!t]
\centering
    \includegraphics[width=0.85\linewidth]{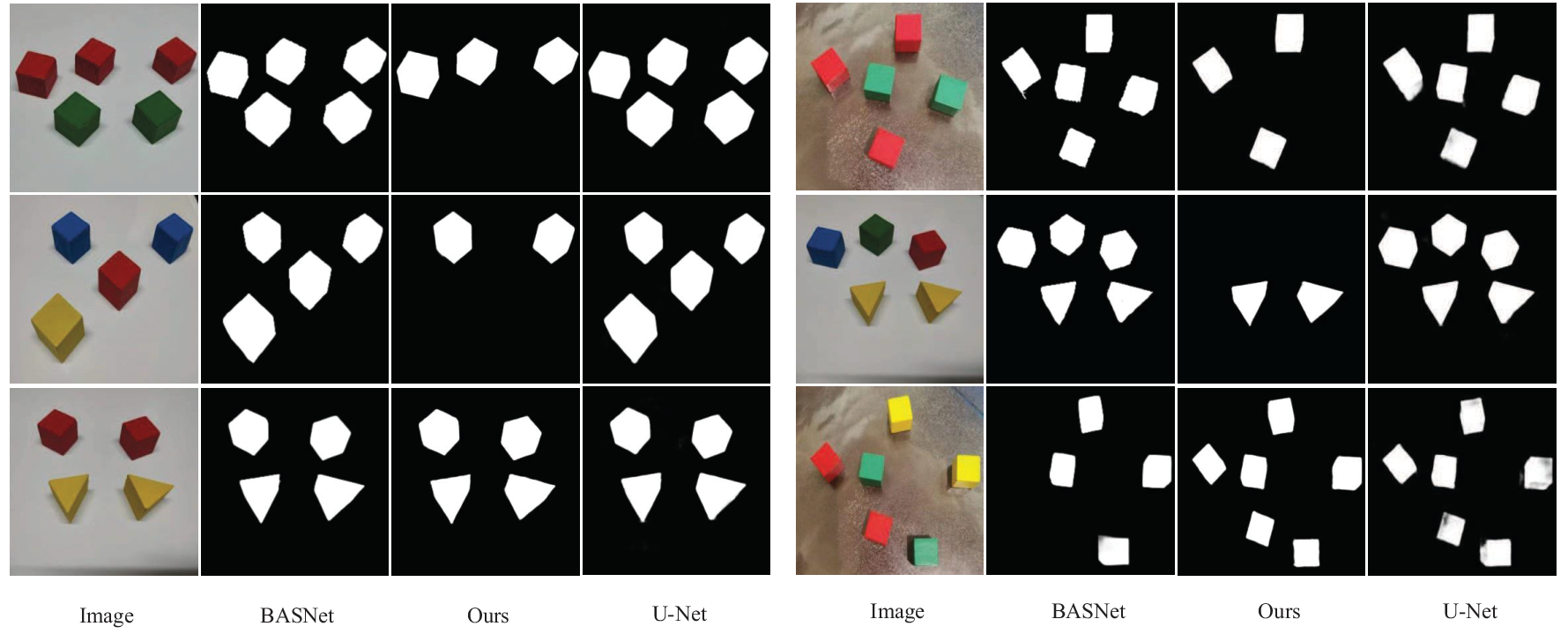}
\caption{Comparison of co-saliency maps generated by U-Net \cite{simonyan2014very}, BASNet\cite{qin2019basnet} and Ours on ambiguity images.}
\label{fig:vis-color}
\end{figure*}

\textcolor{black}{Figure~\ref{fig:vis-color} shows some visual examples of co-saliency maps for ambiguous images. We test some images with ambiguous interpretations in simple scenes to better analyze and understand the role of the RFM module between comparison blocks. The images in the first row of Figure 8 both have two interfering objects with the same color and the same shape. In this situation, the RFM module can select the objects with the highest co-occurrence to be the co-salient objects. The images in the second row have a similar scene, but the interfering objects have different colors and quantities. The RFM module selects the objects with higher co-occurrence and the same color as the co-salient objects. There are many groups of objects with the same number and the same shape and color in the third row. Consequently, there are no domain objects. It is clear that the RFM module is looking for something with the same color or texture, not shape. The reason is that the selected salient region cannot cover the whole object because the positive sample pair augmented the strategy used during the training process. As a result, the model will be unable to take the shape of the object as a factor in the optimization direction.}

\subsection{Comparison with State-of-the-Art Methods}
The results of our method were compared to those of other representative saliency detectors.
Among them, DCL \cite{li2016deep}, RFCN \cite{wang2016saliency}, LFM \cite{Yu2018CoSaliencyDW}, PiCANet \cite{liu2018picanet}, U-Net \cite{simonyan2014very},GCPANet\cite{chen2020global},MINet\cite{pang2020multi} ,BASNet\cite{qin2019basnet},CADC\cite{zhang2021summarize}, GCoNet\cite{fan2021group} and DCFM\cite{yu2022democracy} are deep learning-based methods, CWS \cite{fu2013cluster} and FT \cite{achanta2009frequency} are traditional saliency detection methods. Among the deep learning-based methods, U-Net is a semantic segmentation method; DCL, RFCN, LFM, PiCANet, GCPANet, MINet and BASNet, CADC, GCoNet, and DCFM are saliency detection methods.

\begin{table*}[!t]
\centering {
\caption{The performance of the proposed method and the comparison methods on two datasets. The top three scores are marked in red, blue, and green, respectively. }  \label{tab2}
\scriptsize
\setlength{\tabcolsep}{0.5mm}{
\begin{tabular}{c||cccccc|cccccc}
\Xhline{1.1pt}
\multicolumn{1}{c||}{\multirow{2}{*}{Method}} & \multicolumn{6}{c|}{WhuCoS Dataset - Testset I} & \multicolumn{6}{c}{SDCS Dataset}  \\ \cline{2-13}
\multicolumn{1}{c||}{}  & F-Measure              & Recall                 & Precision              & MAE    &S-measure   &E-measure                & F-Measure     & Recall         & Precision      & MAE   &S-measure   &E-measure         \\ \Xhline{1.1pt}
FT \cite{achanta2009frequency}  & 0.536          & 0.640          & 0.461          & 0.258     &0.088 &0.722     & 0.58           & 0.530          & 0.6            & 0.244  &0.097 &0.692        \\ \hline

CWS \cite{fu2013cluster}  & 0.707          & 0.764          & 0.658          & 0.166   &0.132 &0.743       & 0.767          & 0.7            & 0.79           & 0.165       &0.16  &0.705   \\ \hline

U-Net \cite{ronneberger2015u}  & 0.824          & 0.805          & 0.846          & 0.096   &0.806 &0.934       & 0.856          & 0.762          & 0.889          & 0.107    &0.772  &0.851      \\ \hline
RFCN \cite{wang2016saliency}  & 0.852          & 0.858          & 0.847          & 0.083     &0.819   &0.911     & 0.883          & 0.848          & 0.895          & 0.083    &0.819  &0.911      \\ \hline
PiCANet \cite{liu2018picanet}  & 0.856          & 0.883          & 0.83          & 0.086     &0.791   &0.923     & 0.854          & 0.841          & 0.869          & 0.088      &0.806  &0.904    \\ \hline
DCL \cite{li2016deep} & 0.887          & 0.860          & 0.916          & 0.059   &0.813   &0.876       & 0.888          & 0.844          & 0.902          & 0.059     &0.803  &0.907     \\ \hline

LFM  \cite{Yu2018CoSaliencyDW}                & 0.893              & 0.881              & 0.907              &0.051   &0.765  &0.899
& {\color{green}{\textbf{0.903}}}          & 0.85           & 0.912          & 0.05     &0.79  &0.875     \\ \hline

GCPANet \cite{chen2020global}                & 0.878              & 0.856             & 0.902              &0.058   &0.826 &0.898              & 0.882          & 0.842           & 0.925          & 0.055      &0.834   &0.915 \\ \hline
MINet \cite{pang2020multi}                & 0.89              & 0.870            & 0.911              & 0.052 & 0.823  & 0.928                    & 0.892          & 0.854           & {\color{blue}{\textbf{0.933}}}          & 0.053      &0.852   &0.92    \\ \hline

BASNet \cite{qin2019basnet}                & 0.896              & 0.879              & {\color{green}{\textbf{0.914 }}}            &0.049  &0.862  &0.936                & 0.891          & 0.851           & {\color{red}{\textbf{0.937}}}          & 0.051  &0.857  &0.923          \\ \hline

CADC \cite{zhang2021summarize}               & 0.896              & {\color{blue}{\textbf{0.909}}}             & 0.883              &0.048   &{\color{green}{\textbf{0.884}}} &0.921              & 0.896          & {\color{red}{\textbf{0.887}}}          & 0.905          & {\color{blue}{\textbf{0.047}}}      &{\color{red}{\textbf{0.884}}}   &{\color{red}{\textbf{0.935}}} \\ \hline
GCoNet \cite{fan2021group}                & {\color{blue}{\textbf{0.901}}}            & {\color{green}{\textbf{0.905}}}             & 0.897              & {\color{red}{\textbf{0.047}}} & {\color{blue}{\textbf{0.892}}}  & {\color{green}{\textbf{0.936}}}                  & 0.886          & 0.861           & 0.912          & 0.049      &{\color{red}{\textbf{0.872}}}  &0.926    \\ \hline

DCFM \cite{yu2022democracy}                & {\color{green}{\textbf{0.899}}}             & 0.881              & {\color{red}{\textbf{0.918 }}}            &{\color{red}{\textbf{0.047}}}  &{\color{red}{\textbf{0.895}}}  &{\color{blue}{\textbf{0.943}}}             & 0.894          &{\color{blue}{\textbf{0.881}}}  & 0.908          & {\color{green}{\textbf{0.048}}}  &{\color{red}{\textbf{0.868}}}  &0.928          \\ \hline

Ours (offline)          & 0.892          & 0.870          & {\color{blue}{\textbf{0.917 }}}         & 0.050   &0.839  &0.934     & {\color{blue}{\textbf{0.907}}}          & 0.855          & 0.924         & {\color{green}{\textbf{0.048}}}      &0.821   &{\color{green}{\textbf{0.932}}}   \\ \hline
Ours (online)            & {\color{red}{\textbf{0.904}}} & {\color{red}{\textbf{0.917}}} & 0.892 & {\color{red}{\textbf{0.047}}} &0.851  &{\color{red}{\textbf{0.945}}}  & {\color{red}{\textbf{0.911}}} & {\color{green}{\textbf{0.865}}} & {\color{green}{\textbf{0.925}}} & {\color{red}{\textbf{0.044}}} &0.848   &{\color{red}{\textbf{0.941}}}\\ \Xhline{1.1pt}
\end{tabular}}}
\end{table*}

\begin{table*}[!t]
 \centering {
 \caption{Comparison of the performance on WhuCoS Dataset - Testset II. The top three scores are marked in red, blue, and green, respectively.} \label{tab3}
 \small
 \setlength{\tabcolsep}{0.5mm}{
 {
  \begin{tabular}{c|c||cccccc}
   \Xhline{1.1pt}
   \multirow{1}{*}{Methods} & Venue & F-Measure    & Recall  & Precision   & MAE  &S-measure &E-measure \\
   \Xhline{1.1pt}
   \multirow{1}{*}{FT \cite{achanta2009frequency}} & CVPR'09 & 0.418 & 0.547 & 0.339 & 0.361 & 0.063 &0.672   \\
   \multirow{1}{*}{CWS \cite{fu2013cluster}}  & TIP'13 & 0.664 & 0.709 & 0.626 & 0.213  & 0.106 &0.691  \\
   \multirow{1}{*}{U-Net \cite{ronneberger2015u}}  & MICCAI'15 & 0.788 & 0.733 & 0.851 & 0.129  & 0.742 &0.879   \\
   \multirow{1}{*}{RFCN \cite{wang2016saliency}}  & ECCV'16 & 0.827 & 0.781 & 0.879 & 0.097 & 0.815 &0.902   \\
   \multirow{1}{*}{PiCANet \cite{liu2018picanet}}  & CVPR'18 & 0.822 & 0.80 & 0.845 & 0.092  & 0.793 &0.884  \\
   \multirow{1}{*}{DCL \cite{li2016deep}}  & CVPR'16 & 0.841 & 0.816 & 0.87 & 0.08  & 0.82 &0.869   \\
   \multirow{1}{*}{LFM  \cite{Yu2018CoSaliencyDW}}  & AAAI'18 & 0.892 & {\color{green}{\textbf{0.878}}} & 0.906 & 0.052  & 0.786 &0.913   \\
   \multirow{1}{*}{BASNet \cite{qin2019basnet}}  & CVPR'19 & 0.874 & 0.835 & {\color{blue}{\textbf{0.916}}} & 0.062  & 0.844 &0.926  \\
   \multirow{1}{*}{GCPANet \cite{chen2020global}}  & AAAI'20 & 0.855 & 0.826 & 0.886 & 0.076  & 0.824 &0.891   \\
   \multirow{1}{*}{MINet \cite{pang2020multi}}  & CVPR'20 & 0.860 & 0.821 & 0.903 & 0.074  & 0.832 &0.908   \\
   \multirow{1}{*}{CADC \cite{zhang2021summarize}}  & ICCV'21 & 0.876 & 0.861 & 0.892 & 0.057  & 0.866 &0.924  \\
   \multirow{1}{*}{GCoNet \cite{fan2021group}}  & CVPR'21 & 0.879 & 0.855 & 0.905 & 0.055  & {\color{red}{\textbf{0.874}}} &0.921   \\
   \multirow{1}{*}{DCFM \cite{yu2022democracy}}  & CVPR'22 & {\color{green}{\textbf{0.896}}} & 0.869 & {\color{red}{\textbf{0.924}}} & {\color{green}{\textbf{0.054}}}  & {\color{blue}{\textbf{0.872}}} &{\color{red}{\textbf{0.937}}}   \\
   \hline
   \multirow{1}{*}{Ours (offline) }  &  & {\color{blue}{\textbf{0.903}}} & {\color{blue}{\textbf{0.894}}} & {\color{green}{\textbf{0.912}}} & {\color{blue}{\textbf{0.048}}}  & 0.839 &{\color{green}{\textbf{0.927}}}   \\
   \multirow{1}{*}{Ours (online) }  &  & {\color{red}{\textbf{0.908}}} & {\color{red}{\textbf{0.915}}} & 0.901 & {\color{red}{\textbf{0.046}}}  & {\color{green}{\textbf{0.856}}} &{\color{blue}{\textbf{0.931}}}   \\
   \Xhline{1.1pt}
 \end{tabular}}}}
\end{table*}

\begin{figure*}[t!]
\centering
\includegraphics[width=1.0\linewidth]{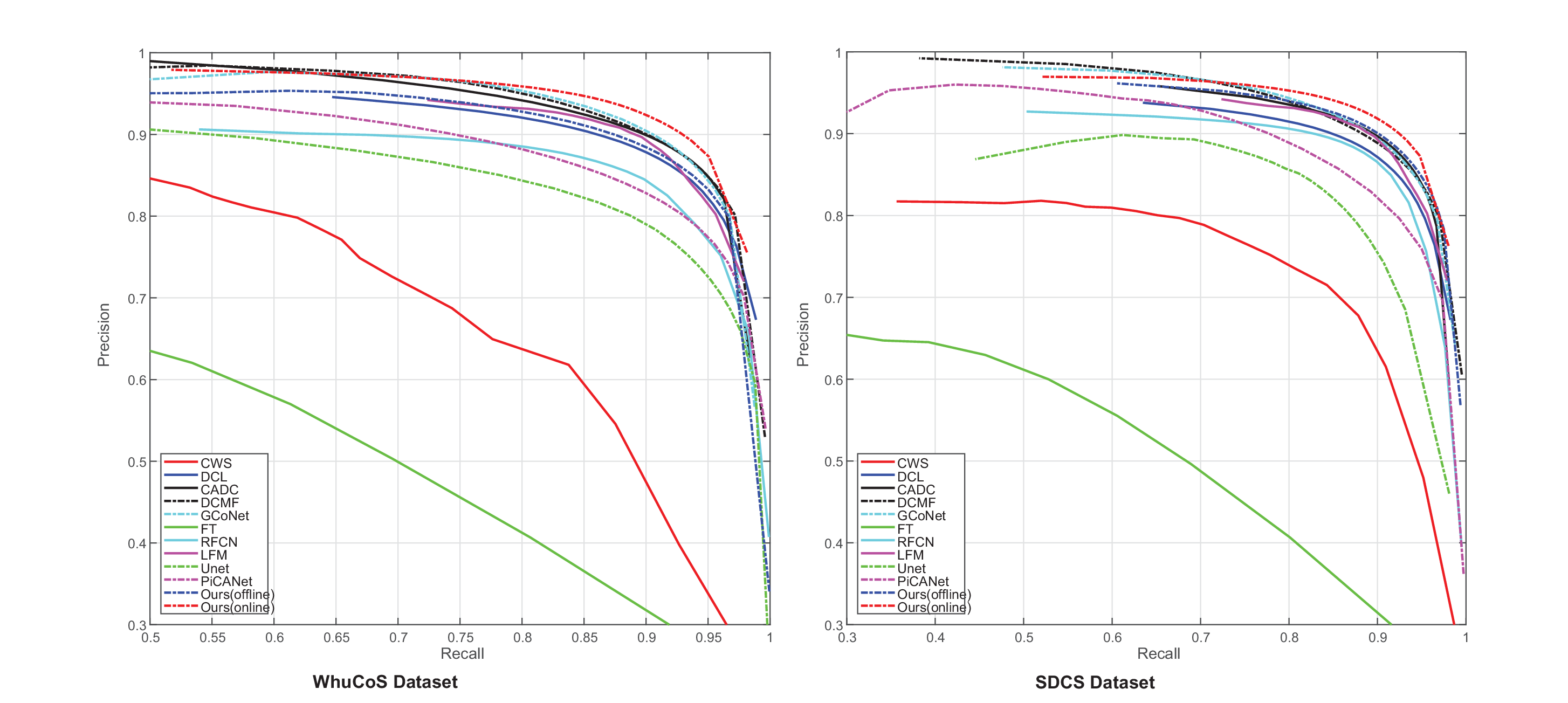}
\caption{Precision-recall curves obtained by the proposed method and the comparison methods on the WhuCoS Testset-I dataset and the SDCS dataset.}
\label{fig:pr}
\end{figure*}

The comparison methods were trained on our dataset by using the recommended parameter settings to ensure a fair comparison.
We compute the average performance for these models with respect to \textcolor{black}{F-measure, S-measure, E-measure, MAE, recall, and precision}. The results are shown in Table~\ref{tab2}.

As shown in Table~\ref{tab2}, deep learning-based methods obtain better saliency results than traditional methods. The baseline model (U-Net) obtains worse saliency results than the other deep learning-based methods.
Our method, which builds an evaluation part on the skip-layer connection structure of U-Net, achieves higher average recall and better average precision than other saliency methods such as DCL, RFCN, PiCANet, GCPANet, and MINet. This is because these saliency detection methods cannot accurately judge co-saliency objects even though trained with the correct co-saliency masks. The accuracy of boundary pixels is slightly lower when compared to the SOTA method, which focuses on boundary discrimination, such as BASNet, but the overall performance is better. Furthermore, as an end-to-end framework, our method has a faster speed than the LFM and a higher average precision. \textcolor{black}{Compared with the SOTA methods in co-saliency detection, e.g., DCFM, CADC, GCoNet, etc., our model achieved higher performance in most metrics in the within-image co-saliency detection task. The possible reason is that the proposed network can make better use of the consensus between objects in a single image.}

For quantitative comparison, the precision-recall curves obtained on the SDCS dataset are plotted in Figure~\ref{fig:pr}. It can be observed that the proposed method has the best performance. Then, the online version of the proposed method, which uses an online training strategy, outperforms the offline version, which is consistent with the results in Table~\ref{tab2}. We also evaluate the performance of these methods on Testset II of the WhuCoS dataset. Table~\ref{tab3} shows the results. We can see that the proposed method obtained a higher performance over the comparison methods that is more significant than that on Testset I. As Testset II is more challenging than Testset I, it simply indicates that the proposed method is better at detecting the co-saliency of challenge samples.

\subsection{Visual comparison}
\begin{figure*}[!t]
\centering
\includegraphics[width=0.9\linewidth]{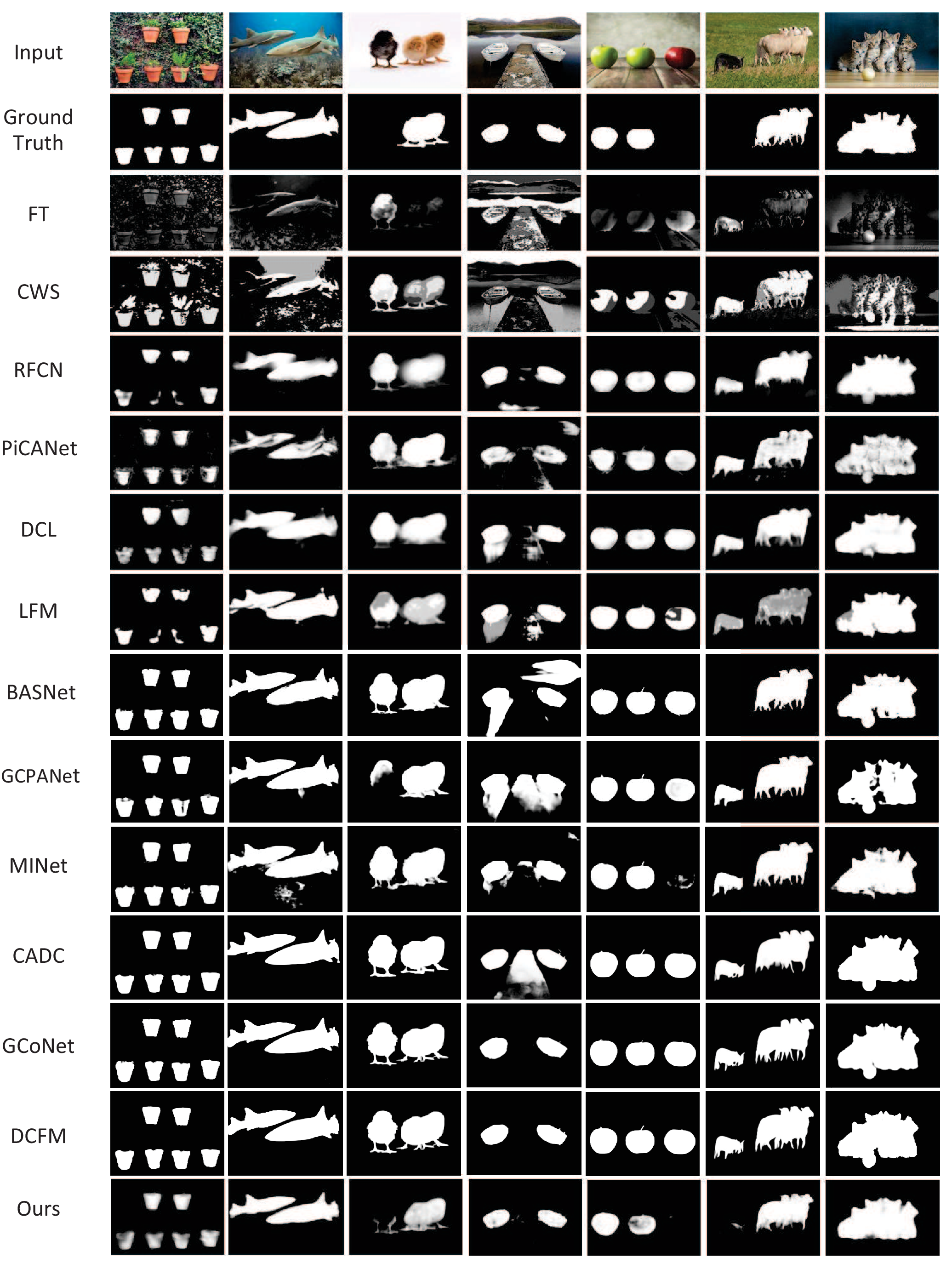}
\caption{Visual comparison of the results obtained by different methods.}
\label{fig:vis-comp}
\end{figure*}

We also visually compare the results obtained by our method and five comparison methods on several sample images, which are shown in Figure~\ref{fig:vis-comp}. Some images are simple, while others are challenging. According to the results, the U-Net structure can save many salient details, which are helpful in producing more accurate edges in the segmentation task. When compared with other methods in hard images, general saliency detectors based on deep learning cannot judge co-saliency objects using only masks as the supervision information. The high-level hidden layers were trained to learn the mapping of the pure saliency sensitivity to co-saliency objects sensitivity, which leads to better results. Figure~\ref{fig:vis-w/o} visually shows the role of RFM in our method.

\section{Conclusion}
In this study, a unified end-to-end network for within-image co-saliency detection was proposed. It combined both top-down and bottom-up strategies. Particularly, an encoder-decoder net was used for co-saliency map prediction in a top-down manner, and an RPN and an RFM were used to guide the model to be sensitive to co-salient regions in a bottom-up manner. An online training sample selection algorithm was presented to enhance the performance of the proposed network. A new dataset WhuCoS was constructed for within-image co-saliency detection. It contains 2,019 natural images, over 300 categories of daily necessities, and 7,000 salient object instances. The experimental results on two target datasets showed that the proposed method achieved state-of-the-art accuracy while running at a speed of 28 fps, and the ablation study validated the effectiveness of the RPN and RFM modules.

\section*{Acknowledgments}
This research was supported by the Key Research and Development Program of Hubei Province under grant No. 2020BAB018, the National Natural Science Foundation of China under grant No.~62171324 and No.~61872277, and the National Key R\&D Program of China under grant No. 2022YFF0901902.

{\small
\bibliographystyle{ieee_fullname}
\bibliography{aaai20}
}

\end{document}